\documentclass[conference]{IEEEtran}
\IEEEoverridecommandlockouts

\usepackage{cite}
\usepackage{amsmath,amssymb,amsfonts}
 
\usepackage{graphicx}
\usepackage{textcomp}
\usepackage{xcolor}
\def\BibTeX{{\rm B\kern-.05em{\sc i\kern-.025em b}\kern-.08em
    T\kern-.1667em\lower.7ex\hbox{E}\kern-.125emX}}

\usepackage{algorithm}
\usepackage{algorithmic}
\usepackage{amsmath}
\usepackage{algorithm}
\usepackage{booktabs}
\usepackage{subcaption}
\usepackage{caption}
\usepackage{enumitem}
\usepackage[normalem]{ulem}
\usepackage{bm}
\usepackage{booktabs}
\usepackage{tikz}
\usepackage{xcolor}
\usepackage{amsthm}
\usepackage{hyperref}

\newtheorem{definition}{Definition}
\newtheorem{problem}{Problem}

\newtheorem{assumption}{Assumption}

\newtheorem{theorem}{Theorem}

\usepackage{amsthm}
\usepackage{enumitem}
\usepackage{hyperref}
\usepackage{url}
\usepackage{multirow}

\newcommand\ci{\perp\!\!\!\perp}
\newcommand\nci{\not \! \perp\!\!\!\perp}

\newcommand{\eat}[1]{}

\begin{document}

\title{Leaning Time-Varying  Instruments for Identifying Causal Effects in Time-Series Data
}

\author{\IEEEauthorblockN{Debo Cheng}
\IEEEauthorblockA{
\textit{STEM}\\
\textit{University of South Australia}\\
Adelaide, Australia \\
debo.cheng@unisa.edu.au}
\and
\IEEEauthorblockN{Ziqi Xu\textsuperscript{*}\thanks{\textsuperscript{*}~Corresponding Authors.}}
\IEEEauthorblockA{
\textit{School of Computing Technologies}\\
\textit{RMIT University}\\
Melbourne, Australia  \\
ziqi.xu@rmit.edu.au}
\and
\IEEEauthorblockN{Jiuyong Li, Lin Liu\textsuperscript{*}, Thuc duy Le}
\IEEEauthorblockA{
\textit{STEM}\\
\textit{University of South Australia}\\
Adelaide, Australia \\
\{jiuyong.li, lin.liu, thuc.duy\}@unisa.edu.au}
\and
\IEEEauthorblockN{Xudong Guo}
\IEEEauthorblockA{
\textit{STEM} \\
\textit{University of South Australia}\\
Adelaide, Australia \\
xudong.guo@mymail.unisa.edu.au}
\and
\IEEEauthorblockN{Shichao Zhang\textsuperscript{*}}
\IEEEauthorblockA{
Guangxi Key Lab of Multi-source Information \\ Mining \& Security,
Guangxi Normal University\\
	Guilin, Guangxi, China \\
 zhangsc@mailbox.gxnu.edu.cn}
}

\maketitle
 
\begin{abstract}
Querying causal effects from time-series data is important across various fields, including healthcare, economics, climate science, and epidemiology. However, this task becomes complex in the existence of time-varying latent confounders, which affect both treatment and outcome variables over time and can introduce bias in causal effect estimation. Traditional instrumental variable (IV) methods are limited in addressing such complexities due to the need for predefined IVs or strong assumptions that do not hold in dynamic settings. To tackle these issues, we develop a novel \underline{T}ime-varying \underline{C}onditional \underline{I}nstrumental \underline{V}ariables (CIV) for \underline{D}ebiasing causal effect estimation, referred to as TDCIV. TDCIV leverages Long Short-Term Memory (LSTM) and Variational Autoencoder (VAE) models to disentangle and learn the representations of time-varying CIV and its conditioning set from proxy variables without prior knowledge. Under the assumptions of the Markov property and availability of proxy variables, we theoretically establish the validity of these learned representations for addressing the biases from time-varying latent confounders, thus enabling accurate causal effect estimation. Our proposed TDCIV is the first to effectively learn time-varying CIV and its associated conditioning set without relying on domain-specific knowledge. Extensive experiments conducted on both synthetic and real-world climate datasets showcase TDCIV’s superior performance in estimating causal effects from time-series data with time-varying latent confounders.
\end{abstract}


\section{Introduction}
\label{sec:intro}
Estimating time-varying causal effects has attracted significant attention across various fields, such as healthcare~\cite{connors1996outcomes}, economics~\cite{grawe2006lifecycle},  climate science~\cite{uppala2005era, zolina2013changes}, and epidemiology~\cite{hernan2006instruments}. 
For example, evaluating the causal effects of a treatment on patient outcomes over time is crucial for developing treatment guidelines in medical contexts. The gold standard for estimating causal effects is to conduct a randomised controlled trial (RCT). However, RCTs are often impossible due to  high costs or ethical concerns~\cite{robins1997causal,imbens2015causal,cheng2024data}. Therefore,  causal effect estimation using observational data has become widely accepted in real-world applications.

\begin{figure}
    \centering
    \includegraphics[width=1\linewidth]{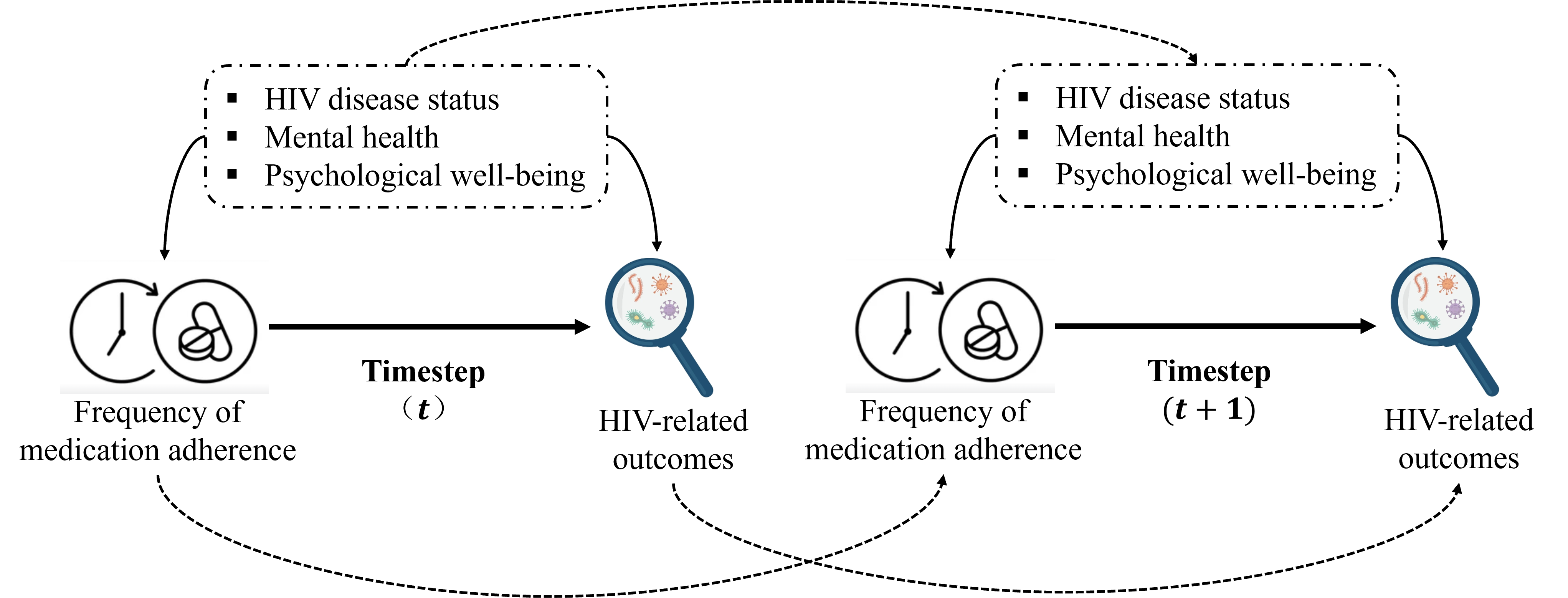}
    \caption{An example of the Women’s Interagency HIV Study demonstrating the impact of treatment adherence on HIV-related outcomes over time. Unmeasured variables, such as HIV disease status, mental health, and psychological well-being, serve as latent confounders that affect both the frequency of medication adherence and HIV viral load. }
    \label{fig:intro}
\end{figure}

Few methods focus on estimating time-varying causal effects from time-series data, largely due to the inherent complexity of the dynamic interactions between variables over time~\cite{hernan2000marginal,bica2019estimating,bica2020time,cui2023instrumental,michael2023instrumental,frauen2023estimating,cao2023estimating,melnychuk2022causal,cheng2024instrumental}. In many real-world applications, estimating time-varying causal effects makes it possible to understand how a disease or policy evolves under different treatments or interventions over time~\cite{bica2020time}. For instance, as shown in Figure~\ref{fig:intro}, medical researchers analyse the time-varying causal effects of treatment adherence on HIV-related outcomes in the Women’s Interagency HIV Study~\cite{michael2023instrumental,cui2023instrumental}. These analyses are essential for designing targeted interventions, optimizing treatment schedules, and improving long-term outcomes for individuals living with HIV.

The main obstacle in estimating time-varying causal effects from time-series data lies in handling time-varying confounders, which dynamically influence both the treatment and the outcome over time, particularly when time-varying latent confounders are also present~\cite{robins1997causal, robins2000marginal}. Neglecting to account for latent confounders can produce biased causal estimates, potentially leading to flawed conclusions and misguided decisions. Following the example above of the Women’s Interagency HIV Study, variables such as HIV disease status, mental health, and psychological well-being are often unmeasured (i.e., latent confounders) and influence both treatment adherence and outcomes over time. Consequently, the causal effect of treatment adherence on HIV-related outcomes becomes unidentifiable due to the causal ambiguities caused by these latent confounders~\cite{pearl2009causality,van2019separators}.

In static settings, the instrumental variable (IV)  approach is commonly used to deal with the latent confounding bias due to latent confounders in causal effect estimation~\cite{hartford2017deep,cheng2023causal}, but to date, few IV methods have been developed for time-series data~\cite{michael2023instrumental,cui2023instrumental,cheng2024instrumental}.
In a static setting, a valid IV must satisfy three conditions: (i) being correlated with the treatment, (ii) being independent of all latent confounders, and (iii) influencing the outcome only through its effect on the treatment~\cite{angrist1995two,brito2002generalized,hernan2006instruments}. The last two conditions are not directly testable, so many existing IV methods rely on predefined IVs, typically chosen by experts based on domain knowledge. This reliance significantly limits the scope of IV applications. To mitigate this limitation, some approaches have been proposed to decompose or recover the information of IVs from observed variables~\cite{yuan2022auto,cheng2023causal}; but assuming static settings. 

In time-series data, identifying a valid time-varying IV from observed variables is more complex and requires stronger assumptions than in the static setting due to the intricate temporal relationships. Thus, most existing IV methods for time-series data require a predefined time-varying IV. For instance, Michael et al.~\cite{michael2023instrumental} developed a weighted estimator with a known time-varying IV to evaluate the effect of hospital type on neonatal survival.  Cui et al.~\cite{cui2023instrumental} extended the standard inverse probability of treatment weighted (IPTW) estimation by identifying parameters of the marginal structural Cox model using a given time-varying IV. Recently, Cheng et al.~\cite{cheng2024instrumental} proposed a Time-varying IV Factor Model (TIFM) to learn the representation of an IV without relying on a predefined time-varying IV, with the assumption that a time-varying IV (though unobserved) exists and influences all time-varying covariates at each time step. However, these IV methods do not account for time-varying Conditional IV (CIV), where a variable serves as a valid IV when conditioned on a set of variables (a conditioning set), a general scenario in many applications. Consequently, learning a time-varying CIV directly from data remains an unresolved challenge.

In this work, we propose a novel \underline{T}ime-varying \underline{C}onditional \underline{I}nstrumental \underline{V}ariables (CIV) for \underline{D}ebiasing causal effect estimation (TDCIV), to disentangle and learn the latent representations of a time-varying CIV and its conditioning set. With the aid of graphical causal models~\cite{pearl2009causality,peters2017elements}, we propose the  time-varying CIV within the full-time DAG $\mathcal{G_{full}}$. Specifically, TDCIV integrates Long Short-Term Memory (LSTM)~\cite{hochreiter1997long}, Variational Autoencoder (VAE)~\cite{kingma2013auto} and Conditional VAE (CVAE~\cite{sohn2015learning}) models to disentangle and learn the representations of a time-varying CIV and its conditioning set, thereby enabling the identification of causal effects in time-series data. These learned representations are then utilised in a CIV method using two stage least square (2SLS)~\cite{angrist1995two} to identify time-varying causal effects in the presence of latent time-varying  confounders. To the best of our knowledge, TDCIV is the first method capable of directly learning the representations of a time-varying CIV and its conditioning set from time-series data while addressing the complexities introduced by time-varying latent confounders.

Therefore, the main contributions of our work can be summarised as follows:
\begin{itemize}
   \item We develop a novel time-varying CIV method for debiasing causal effect estimation (TDCIV). Our TDCIV method aims to disentangle and learn the representations of time-varying CIV and its conditioning set in order to identify causal effects in time-series data. 
   \item We theoretically prove the validity of the learned representations of the time-varying CIV and its conditioning set for identifying causal effects in time-series data with time-varying latent confounders.
  \item Experiments performed on synthetic and real-world climate data validate the effectiveness of TDCIV in identifying causal effects in time-series data influenced by time-varying latent confounders.
\end{itemize}

  Section~\ref{sec:pre} introduces the notations and definitions for our problem and solutions. Section \ref{sec:tdciv} introduces our proposed TDCIV method, detailing the concept of time-varying CIVs and its conditioning set, and the architecture of the TDCIV model. In Section \ref{sec:exp}, we present the experiments on both synthetic and real-world data  to validate the effectiveness of TDCIV and discuss the results. Section \ref{sec:relatedwork} discusses related work in causal inference for time-series data, highlighting the uniqueness and contributions of our method. Finally, Section \ref{sec:con} concludes the paper and proposes potential future work.

\section{Definitions and Problem Setting}
\label{sec:pre}
\subsection{Notations and Basic Definitions}
In this paper, uppercase letters (e.g., $X$) denote variables, while lowercase letters (e.g., $x$) represent their values. Boldfaced uppercase and lowercase letters (e.g. $\mathbf{X}$ and $\mathbf{x}$) are used to indicate sets of variables and their corresponding values, respectively.

We use $\bar{W}_{t}^{(i)}=(W_1^{i}, \dots, W_T^{i})\in\mathcal{W}_T$, $\bar{\mathbf{X}}_{t}^{(i)}=(\mathbf{X}_{1}^{i}, \dots, \mathbf{X}_{T}^{i})\in\mathcal{X}_{T}$ and $\bar{\mathbf{U}}_{t}^{(i)}=(U_1^{i}, \dots, U_T^{i})\in\mathcal{U}_T$ to denote the time-varying treatment, measured covariates, and latent covariates, respectively, over the time step $t \in\{2, \dots, T\}$. Let $\bar{Y}_{t+1}^{(i)}=(Y_2^{i}, \dots, Y_{T+1}^{i})\in\mathcal{Y}_{T+1}$ indicate the outcome of interest from time steps over the time step $t \in\{2, \dots, T+1\}$. The sample index $i$ is omitted unless explicitly required. At any given time step $t$, we represent $k$ measured covariates as $\mathbf{X}_{t} = [\mathbf{X}_{t1}, \dots, \mathbf{X}_{tp}]$ and  the $k$ latent covariates as $\mathbf{U}_{t} = [\mathbf{U}_{t1}, \dots, \mathbf{U}_{tk}]$.. Let $\mathcal{D}=(\bar{\mathbf{X}}_{T}, \bar{W}_{T}, \bar{Y}_{T+1})$ denote the time-series data, and  $\bar{\mathbf{H}}_t = (\bar{\mathbf{X}}_{t}, \bar{\mathbf{W}}_{t-1}, \bar{Y}_t)$ denote the historical data up to time step $t$. 

\begin{figure}[t]
	\centering
	\begin{subfigure}[b]{0.68\linewidth}
		\centering
		\includegraphics[width=\linewidth]{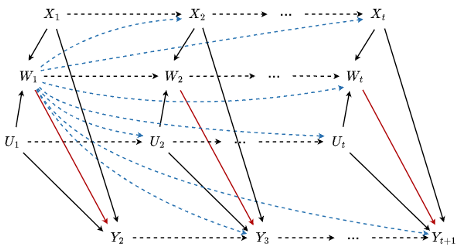}
		\caption{An example full-time causal DAG $\mathcal{G}_{full}$ spanning time $1$ to $t+1$.}
		\label{fig:DAG_full}
	\end{subfigure}
	\hfill
	\begin{subfigure}[b]{0.3\linewidth}
		\centering
		\includegraphics[width=0.715\linewidth]{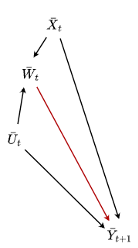}
		\caption{The summary causal DAG $\mathcal{G}$.}
		\label{fig:DAG_summary}
	\end{subfigure}
	\caption{(a) The full-time causal DAG $\mathcal{G}_{full}$ represents causal relationships over time, where $W_{t}$ and $Y_{t+1}$ denote the treatment and outcome, respectively. The time-varying causal effects of $W_{t}$ on $Y_{t+1}$ are shown in red. $\mathbf{X}_t$ and $U_t$ represent time-varying covariates and latent confounders, influencing both $W_{t}$ and $Y_{t+1}$. Blue dashed arrows indicate $W_1$'s effects on subsequent states of all variables, and similar temporal relationships exist for other variables. (b) The summary DAG $\mathcal{G}$ simplifies these mechanisms, omitting latent confounders between covariates $\mathbf{X}_t$ for clarity.}
	\label{fig:DAG01}
\end{figure}

We employ the full-time directed acyclic graph (DAG) $\mathcal{G}_{full} =(\mathbf{V}, \mathbf{E})$~\cite{lauritzen1996graphical,pearl2009causality,peters2017elements,robins2000marginalbook} (i.e., an infinite DAG over time) to represent the causal relationships between variables across temporal dimensions as shown in Fig.~\ref{fig:DAG01} (a), where $\mathbf{V}= \bar{\mathbf{X}}_{t} \cup \bar{\mathbf{U}}_{t}\cup \bar{W}_{t} \cup \bar{Y}_{t+1}$ represents the set of variables, with $\mathbf{E}=\mathbf{V}\times \mathbf{V}$ representing the set of directed edges indicating causal relationships between variables. At each time step $t$, $W_{t}$ directly affects $Y_{t+1}$, represented by the directed edge $W_{t}\rightarrow Y_{t+1}$ in $\mathcal{G}$. In this context, $W_t$ is a parent node of $Y_{t+1}$ in $\mathcal{G}$. Additionally, $X_t$ and $U_t$ each have causal effects on both $W_{t}$ and $Y_{t+1}$, potentially introducing time-varying confouning bias, w.r.t., estimating the causal effects of $W_{t}$ and $Y_{t+1}$ at time step $t$. In the full-time DAG $\mathcal{G}$, $Y_{t+1}$ is a sink node, meaning it has no descendant nodes except its future status $Y_{t+2}$. A summary DAG $\mathcal{G}$ as shown in Fig.~\ref{fig:DAG01} (b) is a simplified representation of the full-time DAG~\cite{peters2017elements}. The summary DAG  $\mathcal{G}$  contains a direct edge between two nodes $V_i$ to $V_j$ if and only if there exists some $1< k$ such that the full-time DAG $\mathcal{G}_{full}$ includes an edge from $V_{it}$ to $V_{j(t+k)}$. 
  
We can read off the conditional independencies in $\mathcal{D}$ from the full-time DAG 
$\mathcal{G}_{full}$ when the Markov property holds. 

\begin{definition}(Markov Property)
A time-series data $\mathcal{D}$ is generated by a full-time DAG $\mathcal{G}_{\text{full}}$, such that $\forall V\in \mathbf{V}$  is conditionally independent of its non-descendants given its parents in $\mathcal{G}_{\text{full}}$. 
\end{definition}
Markov property implies that all dependencies within $\mathcal{D}$ can be derived from the full-time DAG $\mathcal{G}_{\text{full}}$. 

\begin{definition}[Faithfulness]
A time-series dataset $\mathcal{D}$ is faithful to its full-time DAG $\mathcal{G}_{\text{full}}$ if every conditional independence in the joint probability distribution of $\mathcal{D}$ is entailed by the Markov property of $\mathcal{G}_{\text{full}}$, and vice versa. 
\end{definition}
Faithfulness indicates that the full-time DAG $\mathcal{G}_{\text{full}}$ fully encodes all independencies present in $\mathcal{D}$.

Furthermore, we employ a \textit{potential outcomes framework}~\cite{robins1997causal,imbens2015causal} to identify the average causal effects from time-series data. Let $Y_{t+1}(\bar{w}_{t})$ represent the \emph{potential outcomes} associated with the treatment $\bar{w}_t$, where $\bar{w}_{t}$ denotes the treatment applied at time step $t$, just before $Y_{t+1}$ is measured. Not all \emph{potential outcomes} are measured. Using the potential outcomes, we define the average causal effect (ACE) of ${W}_t$ on ${Y}_{t+1}$, denoted as ${ACE}_t(W_t, Y_{t+1})$, from time-series data as follows:
\begin{equation}
\mathbb{E}[Y_{t+1}(\bar{w}_t)\mid \bar{W}_{t-1},\bar{\mathbf{X}}_t] = \mathbb{E}[Y_{t+1}\mid \bar{w}_{t}, \bar{W}_{t-1},\bar{\mathbf{X}}_t]
\label{eq1}
\end{equation}

To identify ${ACE}_t(W_t, Y_{t+1})$ in time-series data, three important assumptions are required~\cite{bica2019estimating,bica2020time,melnychuk2022causal}. The relationship between the observed data and the potential outcomes is established through the \emph{consistency}.
\begin{assumption}[Consistency]
\label{ass:consis}
If $\bar{W}_{t} = \bar{w}_{t}$, then $Y_{t+1}(\bar{w}_{t}) = Y_{t+1}$, i.e., the potential outcome under $\bar{w}_{t}$ is consistent with the observed outcome $Y_{t+1}$.
\end{assumption}

Most existing works~\cite{bica2019estimating,brito2002generalized,Semenova2023Inference,cao2023estimating,sun2023CPT,frauen2023estimating,melnychuk2022causal} rely on the Sequential Randomisation Assumption (SRA) and positivity, introduced by Robins~\cite{robins1997causal,robins2000marginalbook}:

\begin{assumption}[SRA]
    \label{ass:SRA}
    $Y_{t+1}(\bar{w})\ci W_t\mid \bar{W}_{t-1},\bar{\mathbf{X}}_t$, where $\ci$ denotes the statistical independence.
\end{assumption} 
SRA means that there are no latent time-varying confounders, i.e., all variables affecting both ${W}_t$ and $Y_{t+1}(\bar{w})$ are measured and included in the time-series data $\mathcal{D}$. 

\begin{assumption}[Positivity]
	\label{ass:positivity}
If $P(\bar{W}_{t-1} =\bar{w}_{t-1}, \bar{\mathbf{X}}_t = \bar{\mathbf{x}}_t) \neq 0$, then we have $P(\bar{W}_{t} =\bar{w}_{t}\mid \bar{W}_{t-1} =\bar{w}_{t-1}, \bar{\mathbf{X}}_t = \bar{\mathbf{x}}_t) > 0$ for all $w_t$.
\end{assumption} 
The positivity assumption ensures that every possible treatment assignment has a non-zero probability of occurring for any given combination of measured covariates.

\subsection{Problem Definition}

In this work, we consider a more general scenario that involves time-varying latent confounders, denoted by $\bar{\mathbf{U}}_t$, which influences both ${W}_t$ and ${Y}_{t+1}$. Consequently, using Equation~(\ref{eq1}) will result in a biased estimate, as $\bar{\mathbf{U}}_t$ induces a spurious association between ${W}_t$ and ${Y}_{t+1}$ at each time step and cannot be measured directly, i.e., $\mathbb{E}[Y_{t+1}(\bar{w}_{t})\mid \bar{W}_{t-1},\bar{\mathbf{X}}_t] \neq \mathbb{E}[Y_{t+1}\mid \bar{w}_{t}, \bar{W}_{t-1},\bar{\mathbf{X}}_t]$. 

To address the bias caused by time-varying latent confounders $\bar{\mathbf{U}}_t$, we rewrite  Assumption 2 (SRA) by splitting the covariates as measured covariates $\bar{\mathbf{X}}_t$ and latent covariates $\bar{\mathbf{U}}_t$ as Latent SRA (LSRA): 
\begin{assumption}[LSRA]
	\label{ass:LSRA}
	$Y_{t+1}(\bar{w}_{t})\ci W_t\mid \bar{W}_{t-1}, \bar{\mathbf{X}}_t, \bar{\mathbf{U}}_t$.
\end{assumption} 
The assumption states that $\bar{\mathbf{U}}_t$ captures all latent confounding between ${W}_t$ and ${Y}_{t+1}$ such that the SRA holds when conditioning on $\{\bar{\mathbf{X}}_t, \bar{\mathbf{U}}_t\}$. However, $\bar{\mathbf{U}}_t$ is unmeasured and not included in the time-series data $\mathcal{D}$. In our work, we aim to employ the IV-based methods for mitigating the bias caused by $\bar{\mathbf{U}}_t$ in estimating ${ACE}_t(W_t, Y_{t+1})$. 

Most existing IV-based methods for time-series rely on a \textit{given time-varying IV}, which is often impractical in real-world scenarios. In this study, we focus on learning the representations of time-varying CIV $S_t$ and the corresponding conditioning set $\mathbf{Z}_t$  from time-series data. Accordingly, our formal problem setting is defined as:

\begin{problem}
Given a time-series dataset $\mathcal{D}$ of observed variables $\{\bar{\mathbf{X}}_t, \bar{W}_t, \bar{Y}_{t+1}\}$. Assume that $\mathcal{D}$ is faithfule to the underlying full-time DAG $\mathcal{G}_{\text{full}} = (\mathbf{V}, \mathbf{E})$, where $\mathbf{V} = \bar{\mathbf{X}}_{t} \cup \bar{\mathbf{U}}_{t} \cup \bar{W}_{t} \cup \bar{Y}_{t+1}$. $\bar{\mathbf{X}}_t$ represents a set of time-varying measured covariates, while $\bar{\mathbf{U}}_t$ denotes time-varying latent confounders that affect both ${W}_t$ and ${Y}_{t+1}$ across discrete time steps.
Our goal is to estimate $\text{ACE}_t(W_t, Y_{t+1})$ in $\mathcal{D}$, while mitigating the confounding bias introduced by $\bar{\mathbf{U}}_t$.  
\end{problem}

\begin{figure}
    \centering
    \includegraphics[width=.716\linewidth]{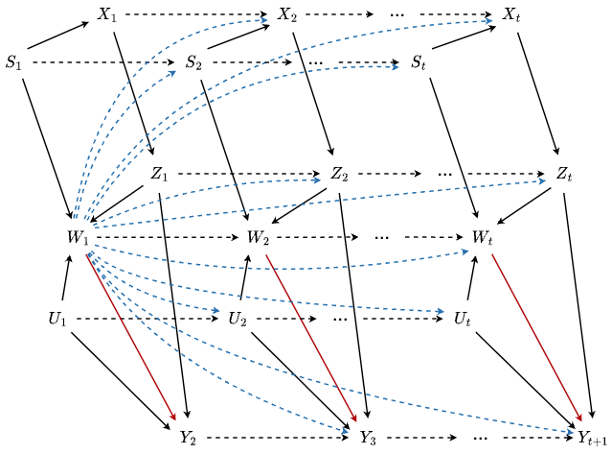}
    \caption{We propose a full-time causal DAG  $\mathcal{G}_{full}$, over time from $1$ to $t$, to disentangle and learn the representations of time-varying CIV and its conditioning set.  Our goal is to query  ${ACE}_t(W_t, Y_{t+1})$ over time, as indicated by the red arrow. $S_t$ and $\mathbf{Z}_t$ are the learned representations of time-varying CIV and its conditioning set. }
    \label{fig:DAG}
\end{figure}

\section{The TDCIV method}
\label{sec:tdciv}
In this section, we introduce the TDCIV to identify causal effects in time-series data where latent confounders evolve over time. First, we present the time-varying CIV, along with its corresponding conditioning set in the full-time DAG $\mathcal{G}_{full}$. Next, we conduct a theoretical analysis of using CIV and representation learning for causal inference in time-series data. Finally, we detail our TDCIV for learning latent representations of a time-varying CIV $S_t$ and the corresponding conditioning set $\mathbf{Z}_t$ using proxy variables in time-series data.

\subsection{Conditional IV (CIV) in a full-time DAG}
\label{subsec:CIV}
The causal effects of  $\text{ACE}_t(W_t, Y_{t+1})$ are not identifiable due to latent confounding bias introduced by $\bar{\mathbf{U}}_t$~\cite{robins2000marginalbook,pearl2009causality,peters2017elements}. IV approach is widely applied for mitigating the latent confounding bias with the aid of a pre-defined IV. However, most IV methods, such as Two-Stage Least Squares (2SLS)~\cite{angrist1995two} and DeepIV~\cite{hartford2017deep}, do not account for the temporal structure of time-series data, leading to biased estimates when directly applied to time-series data to identify causal effects. In this work, we aim to develop a novel time-varying CIV methodology for identifying the causal effects from time-series with time-varying latent confounders. 

First, we extend the concept of CIV from a static causal DAG~\cite{brito2002generalized,cheng2023causal} to a full-time DAG $\mathcal{G}_{full}$, introducing the notion of a time-varying CIV.  In the full-time DAG $\mathcal{G}_{full}$, the time-varying CIV $S_t$ and its corresponding conditioning set $\mathbf{Z}_t$ must satisfy the following three conditions:

\begin{definition}(The concept of a time-varying CIV in a full-time DAG)
\label{def:CIV}
Given a full-time DAG $\mathcal{G}_{full} = (\mathbf{V}, \mathbf{E})$ with $\mathbf{V}= \bar{\mathbf{X}}_{t} \cup \bar{\mathbf{U}}_{t}\cup \bar{W}_{t} \cup \bar{Y}_{t+1}$, and $W_t\rightarrow Y_{t+1}$ in $\mathbf{E}$ at each time-step $t$, a variable $S_t\in \mathbf{X}_t$ is a time-varying CIV relative to $W_t\rightarrow Y_{t+1}$ if there exists a set of variables $\mathbf{Z}_{t}\in \mathbf{X}_t\setminus \{S_t\}$ such that: 
\begin{enumerate}[label=\roman*]
    \item $S_t\nci W_t\mid  \bar{S}_{t-1}, \bar{W}_{t-1} ,\bar{Y}_t, \bar{\mathbf{Z}}_t$, where $\bar{\mathbf{Z}}_t=(\mathbf{Z}_1, \dots, \mathbf{Z}_t)$,
    \item $S_t\ci Y_{t+1}\mid \bar{S}_{t-1}, \bar{W}_{t-1}, \bar{Y}_t, \bar{\mathbf{Z}}_t$ in $\mathcal{G}_{\underline{W_t}}$, where  $\mathcal{G}_{\underline{W_t}}$ is the graph obtained by removing $W_t\rightarrow Y_{t+1}$ at time step $t$ from $\mathcal{G}_{full}$, and 
    \item $\forall Z_t\in\mathbf{Z}_t$ is not a descendant of $Y_{t+1}$.
\end{enumerate}
\end{definition}

The third condition in Definition~\ref{def:CIV} is satisfied in our problem setting because $Y_{t+1}$ is a sink node and has no descendant nodes other than its future statuses (e.g., $Y_{t+2}$ and $Y_{t+3}$). The first condition specifies that $S_t$ is correlated with $W_t$, conditional on the historical values of $\bar{S}_{t-1}$, $\bar{W}_{t-1}$, $\bar{Y}_t$ and $\bar{\mathbf{Z}}_t$. This requirement for a time-varying CIV differs from that of a static CIV, as it necessitates consideration of past states, i.e., the historical data~\cite{peters2017elements,thams2022identifying}. Similarly, the condition (ii) also requires conditioning on the past states of $\bar{S}_{t-1}$, $\bar{W}_{t-1}$, $\bar{Y}_t$ and $\bar{\mathbf{Z}}_t$. We illustrate the time-varying CIV with the full-time DAG as shown in Figure~\ref{fig:DAG}.

To determine a time-varying CIV $S_t$, we need to identify a conditioning set ${\mathbf{Z}}_t$ that, along with the past states of $\bar{S}_{t-1}$, $\bar{W}_{t-1}$, $\bar{Y}_t$ and $\bar{\mathbf{Z}}_{t-1}$,  blocks all paths between $W_t$ and $Y_{t+1}$ in the manipulated causal DAG $\mathcal{G}_{\underline{W_t}}$. Note that the second condition of the time-varying CIV is not testable from time-series data since it is in a manipulated causal graph.

Thus, discovering or determining a valid time-varying CIV  ${S}_t$ and the corresponding conditioning set ${\mathbf{Z}}_t$ from time-series data remains an unsolved problem. In this work, we explore how to learn the time-varying CIV ${S}_t$ and its corresponding conditioning set ${\mathbf{Z}}_t$ directly from time-series data, with as minimal domain knowledge as possible.

To achieve this goal, we extend the concept of proxy variables, originally used in static settings to recover latent variables~\cite{kuroki2014measurement,miao2018identifying}, to time-series data. In causal inference, it is commonly assumed that measurement errors of latent variables can be captured by proxy variables~\cite{louizos2017causal}, thereby facilitating the recovery of the underlying latent variables from these proxies. Building on this concept, we assume that the time-varying CIV is a latent factor but can be approximated through the measurement errors present in the observed covariates, $\bar{\mathbf{X}}_t$, at each time step. Specifically, we assume that at least one proxy variable is available for the time-varying CIV at each step. Additionally, by leveraging the advantages of deep generative models~\cite{kingma2013auto,louizos2017causal,kingma2019introduction}, our objective is to learn sequential representations of time-varying CIV $S_t$ and the corresponding conditioning set $\mathbf{Z}_t$ directly from time-series data with latent time-varying  confounders.

Once we obtain the representations of the time-varying CIV $S_t$ and the corresponding conditioning set $\mathbf{Z}_t$, we can apply the time-varying version of the two-stage least squares (2SLS) estimator to calculate  ${ACE}_t(W_t, Y_{t+1})$ over time  without the latent bias introduced by time-varying confounders~\cite{thams2022identifying}.  

\subsection{The scheme for learning CIV and its conditioning set}
We aim to learn the time-varying CIV ${S}_t$ from proxy variables and generate the conditioning set ${\mathbf{Z}}_t$ based on the observed covariates $\bar{\mathbf{X}}_t$. The time-series data $\mathcal{D}$ is assumed to be generated by an underlying full-time DAG $\mathcal{G}_{full}$, as illustrated in Figure~\ref{fig:DAG}. 
In $\mathcal{G}_{full}$, $S_t$ represents a latent CIV that is proxied by one of the observed variables in $\bar{\mathbf{X}}_t$. The conditioning set ${\mathbf{Z}}_t$ is generated by leveraging $\bar{\mathbf{X}}_t$ along with the historical information. 

 If we have obtained the two representations ${S}_t$  and ${\mathbf{Z}}_t$ from the observed covariates $\bar{\mathbf{X}}_t$, the following theorem ensures that the conditioning set ${\mathbf{Z}}_t$ along with the historical data $\{\bar{S}_{t-1}, \bar{W}_{t-1},\bar{Y}_{t}, \bar{\mathbf{Z}}_{t-1}\}$ instrumentalises ${S}_t$.

\begin{theorem}
\label{theorem:001}
    Given a full-time DAG $\mathcal{G}_{full} = (\mathbf{V}, \mathbf{E})$, where ${\mathbf{V}}= \bar{\mathbf{X}}_{t} \cup \bar{\mathbf{U}}_{t}\cup \bar{W}_{t} \cup \bar{Y}_{t+1}$, and $W_t\rightarrow Y_{t+1}$ in $\mathbf{E}$ at each time-step $t$, and $\bar{\mathbf{U}}_{t}$ representing the set of latent confounders affecting both $W_t$ and $Y_{t+1}$. Suppose that there exists at least one proxy variable of the latent CIV ${S}_t$ in $\bar{\mathbf{X}}_t$. If the representations of ${S}_t$ and ${\mathbf{Z}}_{t}$ as shown in the full-time DAG in Figure~\ref{fig:DAG} can be disentangled and learned from time-series data  $\mathcal{D} = \{\bar{\mathbf{X}}_t, W_t, Y_t\}$, then ${\mathbf{Z}}_t$ along with the historical data $\{\bar{S}_{t-1}, \bar{W}_{t-1},\bar{Y}_{t},\bar{\mathbf{Z}}_{t-1}\}$ instrumentalises ${S}_t$ relative to $W_t\rightarrow Y_{t+1}$ over time.
\end{theorem}
\begin{proof}
    We prove that the representation of $S_t$ is a time-varying CIV when conditioned on the representation ${\mathbf{Z}}_t$ along with the historical data $\{\bar{S}_{t-1}, \bar{W}_{t-1}, \bar{Y}_t, \bar{\mathbf{Z}}_{t-1}\}$ relative to $W_t \rightarrow Y_{t+1}$, as specified by the proposed full-time DAG shown in Figure~\ref{fig:DAG}. 
    
     In the full-time DAG $\mathcal{G}_{full}$, $S_t$ is a cause of $W_t$, meaning that $S_t$ and $W_t$ are dependent given any set of covariates, i.e.,  $S_t\nci W_t\mid  \bar{S}_{t-1}, \bar{W}_{t-1}, \bar{Y}_{t}, \bar{\mathbf{Z}}_t$ holds. Thus, $S_t$ satisfies the first condition of time-varying CIV in Definition~\ref{def:CIV}. Next, in the manipulated causal DAG $\mathcal{G}_{\underline{W_t}}$, all arrows out of $W_t$ are removed. Consequently, the back-door paths between $S_t$ and $Y_{t+1}$ include paths, such as $\dots\rightarrow \mathbf{Z}_t\rightarrow Y_{t+1}$, as well as all past statuses of $\bar{S}_{t-1}$, $\bar{W}_{t-1}$, $\bar{Y}_t$ and $\bar{\mathbf{Z}}_{t-1}$, which are also directed into $Y_{t+1}$. All paths through $\mathbf{Z}_t$ into $Y_{t+1}$ are blocked by $\mathbf{Z}_t$ because these paths end in the sub-path $\mathbf{X}_t \rightarrow \mathbf{Z}_t\rightarrow Y_{t+1}$, forming a chain in $\mathcal{G}_{\underline{W_t}}$. These paths consisting of all past statuses of $\bar{S}_{t-1}$, $\bar{W}_{t-1}$ and $\bar{Y}_t$, which are also directed into $Y_{t+1}$, are blocked by the historical data $\{\bar{S}_{t-1}, \bar{W}_{t-1},\bar{Y}_{t}\}$ since these paths terminated at a direct edge, i.e., $\dots \rightarrow Y_{t+1}$~\cite{thams2022identifying}. Therefore, $S_t$ is d-separated from $Y_{t+1}$ by conditioning on $\bar{S}_{t-1}, \bar{W}_{t-1}, \bar{Y}_t$ and $\bar{\mathbf{Z}}_t$, which blocks all back-door paths between $S_t$ and $Y_{t+1}$. Thus, $S_t\ci Y_{t+1}\mid \bar{S}_{t-1}, \bar{W}_{t-1}, \bar{Y}_t, \bar{\mathbf{Z}}_t$ holds, and the representations of $S_t$ and ${\mathbf{Z}}_t$ satisfy the second condition of time-varying CIV in Definition~\ref{def:CIV}. Finally, since $Y_{t+1}$ is a sink node in $\mathcal{G}$, ${\mathbf{Z}}_t$ does not contain any descendants of $Y_{t+1}$. Thus, the representations of $S_t$ and ${\mathbf{Z}}_t$ satisfy the third condition of time-varying CIV in Definition~\ref{def:CIV}.
    
    Therefore, we conclude that ${\mathbf{Z}}_t$ along with the historical data $\{\bar{S}_{t-1}, \bar{W}_{t-1},\bar{Y}_{t},\bar{\mathbf{Z}}_{t-1}\}$ instrumentalises ${S}_t$ relative to $W_t\rightarrow Y_{t+1}$ for unbiased  $\text{ACE}_t(W_t, Y_{t+1})$ estimation from time-series data with time-varying latent confounders.
\end{proof}

\subsection{The Proposed TDCIV Method}
 Inspiration from the effective use of proxy variables in causal inference~\cite{kuroki2014measurement,louizos2017causal,miao2018identifying,zhang2021treatment,cheng2023causal}, leverage the strengths  of deep generative models~\cite{kingma2013auto,kingma2019introduction} to directly capture the sequential representations of CIV ${S}_t$ and its conditioning set ${\mathbf{Z}}_t$ from time-series data at each time step. We explore the use of VAEs and LSTM networks~\cite{hochreiter1997long} to disentangle and learn latent representations of time-varying CIV and its corresponding conditioning set.

To capture the historical information inherent in $\bar{\mathbf{X}}_t$, $\bar{W}_{t-1}$ and $\bar{Y}_{t}$ (i.e., $\bar{\mathbf{H}}_t$), we utilise LSTM as a core component of the TDCIV method. By integrating LSTM at each time step, the method effectively models the historical dependencies present in time-series data.
To disentangle and learn ${S}_t$ and ${\mathbf{Z}}_t$ from $\bar{\mathbf{X}}_t$ based on the causal DAG as shown in Figure~\ref{fig:DAG}, we also employ VAE and CVAE models. 

\begin{figure*}[t]
	\centering
	\includegraphics[scale=0.405]{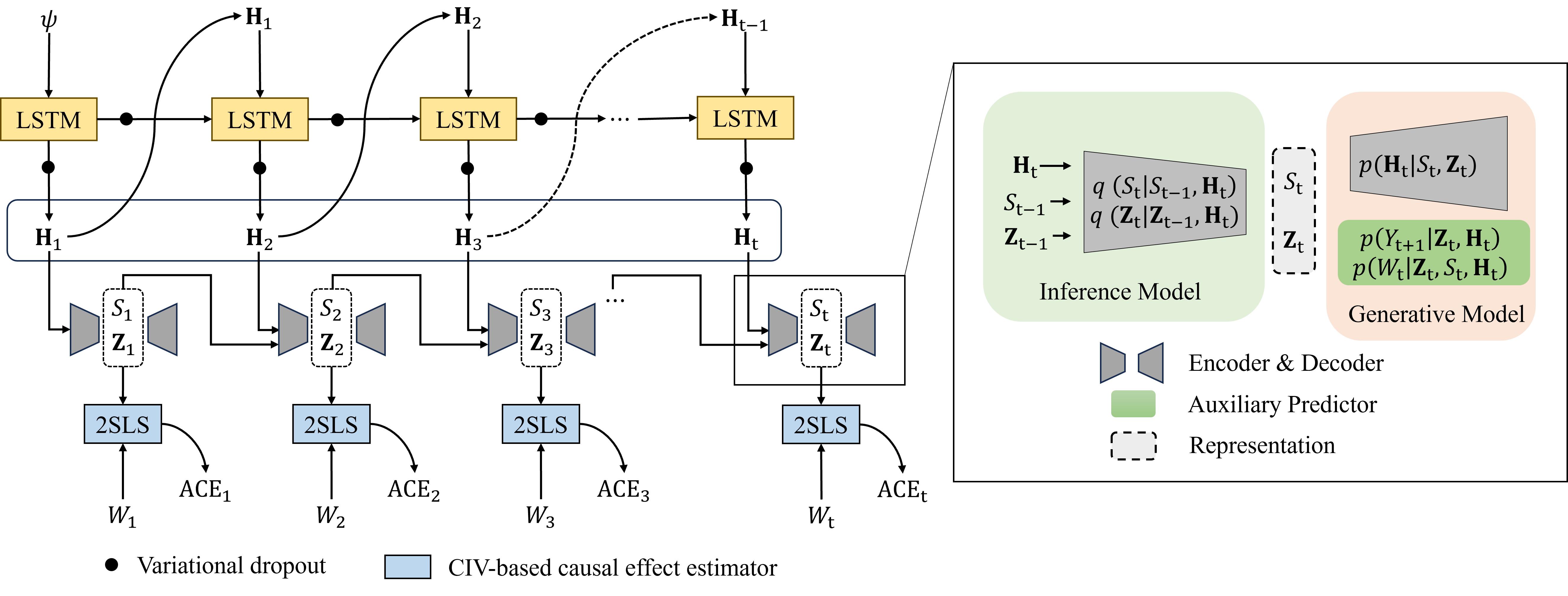}
	\caption{An overview of the  TDCIV's architecture: The LSTM generates the historical data $\mathbf{H}_t$ as a function of the previous historical data $\mathbf{H}_{t-1}$ and the current input. Within the blue rectangle, the 2SLS estimators can be substituted with any CIV-based causal effect estimator. On the right side, the inference and generative networks collaboratively learn the latent representations of the time-varying CIV, $S_t$, and its corresponding conditioning set, $\mathbf{Z}_t$, over time.}
	\label{pic:model}
\end{figure*}

We demonstrate our  architecture of the TDCIV  in Figure~\ref{pic:model}.  Our TDCIV consists of three components: an LSTM, an inference network, and a generative network. Specifically, the inference and generative networks are combined to form the VAE, which approximates the posterior distributions $p(S_t \mid \bar{S}_{t-1}, \bar{\mathbf{H}}_t)$ and $p(\mathbf{Z}_t \mid \bar{\mathbf{Z}}_{t-1}, \bar{\mathbf{H}}_t)$. As depicted in the full-time causal DAG in Figure~\ref{fig:DAG}, these distributions for the two latent representations, $S_t$ and $\mathbf{Z}_t$, at time step $t$ rely on the historical information $\mathbf{H}_t$, as well as their respective previous states, $\bar{S}_{t-1}$ and $\bar{\mathbf{Z}}_{t-1}$. These latent representations capture the time-varying CIV and its conditioning set at time step $t$. 

As discussed previously, we use the LSTM component as a core part of our TDCIV to capture the historical information between $\bar{\mathbf{X}}_t$, $\bar{W}_{t-1}$ and $\bar{Y}_{t}$. The LSTM is defined as follows:
\begin{equation}
	\begin{aligned}
		&\mathbf{H}_{1} = \text{LSTM}( \xi );\\
		&\mathbf{H}_{t} = \text{LSTM}(\mathbf{H}_{t-1}),
	\end{aligned}
 \label{eq:002}
\end{equation} 
where $ \xi$ represents the randomly initialised parameter at the initial step, which is then trained jointly with the other parameters of the LSTM.  Then, $\bar{\mathbf{H}}_t$ is passed forward to subsequent time steps and is used alongside the inference and generative networks to model the latent CIV and its conditioning set. The LSTM's capability to retain long-term associations in the sequence ensures that the historical context is effectively integrated into the learning process of our TDCIV. Upon training the LSTM, we extract $\bar{\mathbf{H}}_t$ and use it as an input for the subsequent representation learning process.

We then discuss how to utilise the components of the VAE model, including the inference network and the generative network, to recover $S_t$ and $\mathbf{Z}_t$ from the time-series data.

In the inference network, two separate encoders $q(\mathbf{Z}_t\mid \bar{\mathbf{H}}_t,\bar{\mathbf{Z}}_{t-1})$ and $q(S_t\mid \bar{\mathbf{H}}_t, \bar{S}_{t-1})$ are employed to approximate the posterior distributions of $S_t$ and $\mathbf{Z}_t$, respectively. In the generative network, we use a single decoder, $p(\bar{\mathbf{H}}_t \mid \mathbf{Z}_{t}, S_t)$ to reconstruct $\bar{\mathbf{H}}_t$. Following the standard VAE structure as described in the literature~\cite{kingma2013auto,kingma2019introduction},  we model the prior distribution $p(S_t)$ using a Gaussian distribution, which can be written as follows:
\begin{equation}
	\label{eq:003}
	\begin{aligned}
		p(S_t) =  \mathcal{N}(0, 1),
	\end{aligned}	
\end{equation}
\noindent where $\mathcal{N}(\cdot)$ denotes the Gaussian distribution. 

Additionally, the encoders $q(\mathbf{Z}_t\mid \bar{\mathbf{H}}_t,\bar{\mathbf{Z}}_{t-1})$ and $q(S_t\mid \bar{\mathbf{H}}_t, \bar{S}_{t-1})$ in the inference network are formulated as follows:
\begin{equation}
	\label{eq:004}
	\begin{aligned}
		q(\mathbf{Z}_t \mid \bar{\mathbf{H}}_t, \bar{\mathbf{Z}}_{t-1}) &= \prod_{i=1}^{D_{\mathbf{Z}_t}} \mathcal{N}(\mu = \hat{\mu}_{\mathbf{Z}_{t_i}}, \sigma^2 = \hat{\sigma}^2_{\mathbf{Z}_{t_i}});\\
		q(S_t \mid \bar{\mathbf{H}}_t, \bar{S}_{t-1}) &= \prod_{i=1}^{D_{S_t}} \mathcal{N}(\mu = \hat{\mu}_{S_{t_i}}, \sigma^2 = \hat{\sigma}^2_{S_{t_i}}),
	\end{aligned}	
\end{equation} 
\noindent where $\hat{\mu}_{\mathbf{Z}_t}, \hat{\mu}_{S_t}$ and $\hat{\sigma}^2_{\mathbf{Z}_t}, \hat{\sigma}^2_{S_t}$ are the means and variances of the functions of $\bar{\mathbf{H}}_t, \bar{\mathbf{Z}}_{t-1}$, and $\bar{S}_{t-1}$, parameterised by neural networks. The dimensions of $\mathbf{Z}_t$ and $S_t$ are denoted by $D_{\mathbf{Z}_t}$ and $D_{S_t}$, respectively.

To generate the distribution $\mathbf{Z}_t$ conditioned on  historical data $\bar{\mathbf{H}}_t$ in our generative model, we follow the standard conditional VAE (CVAE) network~\cite{sohn2015learning}, using Monte Carlo (MC) sampling to generate $\mathbf{Z}_t$ conditioned on historical data $\bar{\mathbf{H}}_t$ as follows:
\begin{equation}
		\label{eq:005}
	\begin{aligned}
		\mathbf{Z}_t \backsim p(\mathbf{Z}_t\mid \bar{\mathbf{H}}_t)
	\end{aligned}	
\end{equation} 

Moreover, in our generative networks, the treatment $W_t$ and $\bar{\mathbf{H}}_t$ at time-step $t$ are generated as follows:
\begin{equation}
		\label{eq:006}
	\begin{aligned}
		p(W_t\mid \bar{\mathbf{Z}}_t, \bar{S}_t, \bar{\mathbf{H}}_{t}) &= Bern(\sigma(g_1(\bar{\mathbf{Z}}_t, \bar{S}_t, \bar{\mathbf{H}}_{t})));\\ 
      p(\bar{\mathbf{H}}_t \mid \mathbf{Z}_{t}, S_t) &= \prod_{i=1}^{D_{\bar{H}_{ti}}} p(\bar{H}_{ti}\mid \mathbf{Z}_t, S_t),
	\end{aligned}	
\end{equation} where $g_1(\cdot)$ represents a function parameterised by neural networks, and $\sigma(\cdot)$ denotes the logistic function.

In the generative network, the outcome $Y_{t+1}$ at time step $t+1$ is generated based on the learned representations  $\bar{\mathbf{Z}}_t$ and $\bar{S}_t$. Furthermore, the historical data $\bar{W}_t$ and $\bar{Y}_t$ are crucial for generating $Y_{t+1}$ at time step $t+1$. The generative process for $Y_{t+1}$ adheres to the CVAE, with $\bar{\mathbf{Z}}_t$, $\bar{S}_t$, and $\bar{\mathbf{H}}_t$ as inputs.

For continuous outcomes, $Y_{t+1}$ is modelled as a Gaussian distribution:
\begin{equation}
\label{e1:007}
\begin{aligned}
p(Y_{t+1} \mid \bar{\mathbf{Z}}_t, \bar{\mathbf{H}}_t)  & = \mathcal{N}(\mu = g_2(\bar{\mathbf{Z}}_t,  \bar{\mathbf{H}}_t), \\ & \sigma^2 = g_3(\bar{\mathbf{Z}}_t, \bar{\mathbf{H}}_t)),	
\end{aligned}
\end{equation}
where $g_2(\cdot)$ and $g_3(\cdot)$ are functions parameterised by neural networks that compute the mean and variance of the Gaussian distribution. For binary outcomes, $Y_{t+1}$ is modelled using a Bernoulli distribution:
\begin{equation}
p(Y_{t+1}\mid \bar{\mathbf{Z}}_t,   \bar{\mathbf{H}}_t) = \text{Bern}(\sigma(g_4(\bar{\mathbf{Z}}_t,  \bar{\mathbf{H}}_t))),
\end{equation}
where $g_4(\cdot)$ is a function parameterised by neural networks. The parameters of the model can be optimised by maximising the evidence lower bound (ELBO). The ELBO is given by:
\begin{equation}
\begin{aligned}
\mathcal{M} & = \mathbb{E}_{q}\left[\log p(\bar{\mathbf{H}}_t \mid \mathbf{Z}_{t}, S_t)\right] \\ & 
- D_\text{KL}(q(S_t \mid \bar{\mathbf{H}}_t, \bar{S}_{t-1}) \parallel p(S_t)) 
\\ &
- D_\text{KL}(q(\mathbf{Z}_t \mid \bar{\mathbf{H}}_t, \bar{\mathbf{Z}}_{t-1}) \parallel p(\mathbf{Z}_t\mid \bar{\mathbf{H}}_t)),
\end{aligned}
\end{equation} where $D_\text{KL}$  is the Kullback–Leibler divergence. 

To ensure that $S_t$ captures as much information as possible about the CIV, and that $\mathbf{Z}_t$ captures more confounding information between $S_t$ and $Y_{t+1}$, we add two predictors into our ELBO to predict $W_t$ and $Y_{t+1}$, respectively, as done in previous work~\cite{louizos2017causal,zhang2021treatment,cheng2023causal}. In our TDCIV,  $W_t$ is predicted by $\bar{\mathbf{Z}}_t$, $\bar{S}_t$, and $\bar{\mathbf{H}}_{t}$, while the outcome $Y_{t+1}$ is predicted by $\bar{\mathbf{Z}}_t$, and $\bar{\mathbf{H}}_{t}$. Thus, our final objective function is expressed as:
\begin{equation}
\begin{aligned}
\mathcal{L}_{TDCIV} &= -\mathcal{M} + \alpha 
\mathbb{E}_{q}\left[\log p(W_t \mid \bar{\mathbf{Z}}_t, \bar{S}_t, \bar{\mathbf{H}}_{t})\right]  \\& + \beta \mathbb{E}_{q}\left[\log p(Y_{t+1} \mid \bar{\mathbf{Z}}_t, \bar{\mathbf{H}}_{t})\right],
\end{aligned}
\end{equation}
where $\alpha$ and $\beta$ are hyper-parameters that balance the two additional predictors.

\begin{algorithm}[t]
\label{alg01}
\caption{TDCIV (Time-varying CIV using Deep generative model)}
\begin{algorithmic}
\STATE \textbf{Input}: Time-series data $\mathcal{D} = \{\bar{\mathbf{X}}_t, \bar{W}_t, \bar{Y}_t\}$ for time steps $t = 1, 2, \dots, T$\\
\STATE \hspace*{0em}\textbf{Output}: Estimated ${ACE}_t(W_t, Y_{t+1})$
\STATE \textbf{Initialise} LSTM model with hidden states $\mathbf{H}_t$ to capture time dependencies in $\mathcal{D}$
\FOR{each time step $t$}
    \STATE \textbf{Compute} hidden state $\mathbf{H}_t$ using Equation~\ref{eq:002};
    \STATE \textbf{Inference Step}: \\
    Use $\mathbf{H}_t$ in VAE and CVAE: $q(\mathbf{Z}_t \mid \bar{\mathbf{H}}_t, \bar{\mathbf{Z}}_{t-1})$ and $q(S_t \mid \bar{\mathbf{H}}_t, \bar{S}_{t-1})$ are employed to learn the representations of the time-varying CIV $S_t$ and its conditioning set $\mathbf{Z}_t$;
    \STATE \textbf{Generate Step}: \\
    Reconstruction distributions: $p(\mathbf{H}_t \mid \bar{\mathbf{Z}}_t, \bar{S}_t)$, $p(W_t \mid \bar{\mathbf{Z}}_t, \bar{S}_t, \bar{\mathbf{H}}_{t})$ and $p(Y_{t+1} \mid \bar{\mathbf{Z}}_t, \bar{\mathbf{H}}_{t})$;
    \STATE   Update parameters of $\mathcal{L}_{TDCIV}$;
    \STATE \textbf{Prediction Step}: \\
    Obtain $S_t$ and $\mathbf{Z}_t$, and use them in 2SLS to estimate the average causal effect ${ACE}_t(W_t, Y_{t+1})$ at time step $t$;
\ENDFOR
\STATE \textbf{Return}: ${ACE}_t(W_t, Y_{t+1})$
\end{algorithmic}
\end{algorithm}

To estimate ${ACE}_t(W_t, Y_{t+1})$, we extract the time-varying representations $S_t$ and $\mathbf{Z}_t$ from our TDCIV model at each time step $t$. The representation $S_t$ captures the information of time-varying CIV, while $\mathbf{Z}_t$, along with $\bar{\mathbf{H}}_t$, forms the conditioning set. Once both representations are obtained, we employ the CIV method~\cite{angrist1995two,thams2022identifying} to estimate ${ACE}_t(W_t, Y_{t+1})$ at time step $t$. In the linear case, ${ACE}_t(W_t, Y_{t+1})$ is calculated as:
\begin{equation}
	ACE_t(W_t,Y_{t+1}) = \frac{\sigma_{S_t*Y_{t+1}*(\bar{\mathbf{Z}}_t,\bar{\mathbf{H}}_t)}}{\sigma_{S_t*W_t*(\bar{\mathbf{Z}}_t,\bar{\mathbf{H}}_t)}},
\end{equation}
where $\sigma_{S_t*Y_{t+1}*(\bar{\mathbf{Z}}_t,\bar{\mathbf{H}}_t)}$ and $\sigma_{S_t*W_t*(\bar{\mathbf{Z}}_t,\bar{\mathbf{H}}_t)}$ represent the estimated causal effect of $S_t$ on $Y_{t+1}$ and on $W_t$, conditioned on $\bar{\mathbf{Z}}_t$ and $\bar{\mathbf{H}}_t$ based on Theorem~\ref{theorem:001}.

We present the pseudo-code for the TDCIV method in Algorithm~1.  In detail, our TDCIV begins by initialising an LSTM  model to capture relevant historical information and encodes the sequential information into $\mathbf{H}_t$. Then,  ${\mathbf{H}}_t$ is used in VAE and CVAE to disentangle and learn the latent representations of time-varying CIV ${S}_t$ and the corresponding conditioning set $\mathbf{Z}_t$. Next, TDCIV uses the learned latent representations to reconstruct the observational time-series data and update the parameters of $\mathcal{L}_{TDCIV}$. Finally, we extract $S_t$ and $\mathbf{Z}_t$ at each time step $t$ and use them in 2SLS to estimate  ${ACE}_t(W_t, Y_{t+1})$. The output of our TDCIV is a list of ${ACE}_t(W_t, Y_{t+1})$ over time.

\textbf{Time Complexity}: The time complexity of  TDCIV  consists of three components: the computational costs associated with the LSTM model, the VAE/CVAE models, and the 2SLS estimation. A detailed analysis of each component is provided below: (1) \textbf{The time complexity of LSTM model}: The LSTM captures temporal dependencies with a LSTM layer that requires \( O(T m^2) \) operations per layer, where $m$ is the number of hidden units. Thus, the total time complexity of the LSTM is  $O(L*T*m^2)$, where $L$ is the number of LSTM layers. (2) \textbf{The time complexity of VAE and CVAE Models}: The VAE and CVAE learn latent representations of  $S_t$ and  $\mathbf{Z}_t$ from time-series data $\mathcal{D}$. The encoding and decoding for each time step have a complexity \( O(\phi ^2 +  \phi\varphi) \), leading to  $O(n \cdot T \cdot (\phi^2 + \phi\varphi))$, where $\varphi$ is the dimensionality of the latent space and $\phi$ is the number of neurons per dense layer. (3) \textbf{The time complexity of 2SLS Estimation}: The 2SLS step estimates the treatment effect ${ACE}_t(W_t, Y_{t+1})$ with complexity $O(n \eta^2)$, where $\eta$ is the number of $\bar{\mathbf{X}}_t$.
 
Therefore,  the total time complexity of TDCIV can be expressed as  $O(n \cdot T \cdot (m^2 + \phi^2 + \phi \varphi) + n \eta^2)$.

The time complexity analysis of TDCIV highlights that its effectiveness is influenced by $T$, $m$, $\phi$, $\varphi$ and $\eta$. This also indicates that the TDCIV method is well-suited for high-dimensional, long-sequence time-series data, particularly in scenarios where time-varying latent confounders are present.

\textbf{Limitations}: The performance of TDCIV relies on the assumption regarding the measurement errors of the time-varying CIV present in $\bar{\mathbf{X}}_t$. If this assumption is violated, TDCIV is likely to produce estimations with high variance. Additionally, he TDCIV requires that the latent representations learned by the VAE and CVAE correct. Failure to adequately learn these representations due to insufficient data, suboptimal hyperparameter settings, or overly complex models, may negatively impact performance. In such scenarios, we recommend that users conduct sensitivity analyses.

\section{Experiments}
\label{sec:exp}

In this section, we examine the effectiveness of TDCIV in accurately estimating  ${ACE}_t(W_t, Y_{t+1})$ in time-series data $\mathcal{D}$. The experiments are divided into two parts: evaluation using synthetic datasets and a real-world case study involving climate data. In the first part, we construct synthetic datasets based on the methodologies proposed in \cite{wang2019blessings, bica2020time}. These datasets provide access to ground truth, enabling precise computation of the true causal effects over time. Additionally, we perform sensitivity analyses to evaluate TDCIV's robustness under various parameter configurations.
In the second part, we validate the practical applicability of TDCIV by applying it to   climate data. This case study demonstrates the method's effectiveness in uncovering causal relationships in complex, real-world scenarios.

\begin{figure*}[t]
	\centering
	\includegraphics[scale=0.423]{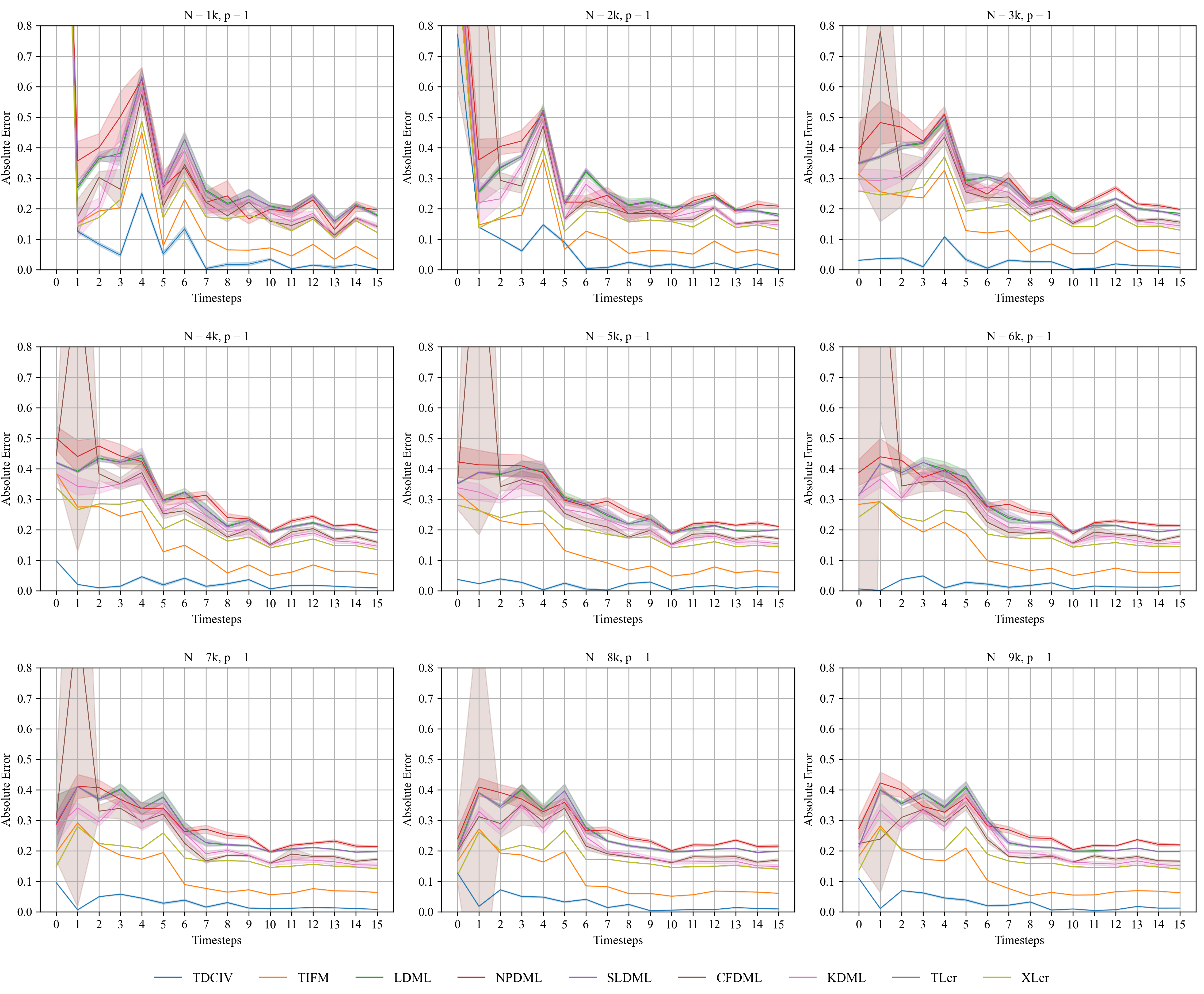}
	\caption{Comparison of absolute errors across all methods, presented with their means and standard deviations calculated over 30 synthetic datasets.}
	\label{fig:res_simulate}
\end{figure*}

\subsection{Experiment Setup}
\paragraph{Data Generation}
To enable fair comparisons, we conduct simulation studies using multiple synthetic datasets. To emulate real-world scenarios, we adopt the approach outlined in~\cite{bica2020time}, employing $p$-order autoregressive processes to generate synthetic data. At each time step $t$, the observed covariates $\mathbf{X}_t$ and latent covariates $\mathbf{U}_t$ are defined as follows:
\begin{equation}
	\begin{aligned}
		\mathbf{X}_{t} &= \frac{1}{p} \sum_{i = 1}^{p} \left( \alpha_i \mathbf{X}_{t-i} + \omega_i W_{t-i} \right) + \varepsilon_{\mathbf{X}}, \\
		\mathbf{U}_{t} &= \frac{1}{p} \sum_{i = 1}^{p} \left( \beta_i \mathbf{U}_{t-i} + \lambda_i W_{t-i} \right) + \varepsilon_{\mathbf{U}}.
	\end{aligned}
	\label{eq:dynamic_system}
\end{equation}

The parameters in Equation \ref{eq:dynamic_system} are sampled as $\alpha_i, \lambda_i \sim \mathcal{N}(0, 0.5^2)$, $\omega_i, \beta_i \sim \mathcal{N}(1 - (i/p), (i/p)^2)$, and noise terms $\varepsilon_{\mathbf{X}}, \varepsilon_{\mathbf{U}} \sim \mathcal{N}(0, 0.01^2)$.
The latent time-varying CIV ${S}_{t}$ is generated as follows:
\begin{equation}
            {S}_{t} = \frac{1}{p} \sum_{i = 1}^{p} ({S}_{t-i}) + \varepsilon_{{S}}
\end{equation}
The conditioning set $\mathbf{Z}_{t}$ is generated as:
\begin{equation}
	\begin{aligned}
	\mathbf{Z}_{t} = \frac{1}{p} \sum_{i = 1}^{p} (\mathbf{Z}_{t-i} + \mathbf{X}_{t}) + \varepsilon_{\mathbf{U}},
	\end{aligned}
\end{equation}
The time-varying treatment variable $W_t$ depends on the latent CIV $S_t$, its conditioning set $\mathbf{Z}_{t}$, latent confounders $\mathbf{U}_t$ and covariates $\mathbf{X}_t$:
\begin{equation}
	\begin{aligned}
		&\theta_t = \mu_{\mathbf{X}}\widehat{\mathbf{X}}_t + \mu_{\mathbf{U}}\widehat{\mathbf{U}}_t + \mu_{{S}}\widehat{{S}}_t + \mu_{\mathbf{Z}}\widehat{\mathbf{Z}}_t,\\
		&W_t \mid \theta_t \sim Bernoulli(\sigma(c \cdot \theta_t)),
	\end{aligned}
\end{equation}
\noindent where $\widehat{\mathbf{X}}_t, \widehat{\mathbf{U}}_t, \widehat{{S}}_t, \widehat{\mathbf{Z}}_t$ denote the sum for the measured covariates, latent covariates, latent CIV, and its conditioning set, respectively, over the last $p$ time steps. The function $\sigma(\cdot)$ denotes the sigmoid function, and $\mu_{\mathbf{X}_t}, \mu_{\mathbf{U}_t}, \mu_{{S}_t}, \mu_{\mathbf{Z}_t}, c \sim \mathcal{N}(0,1^{2})$. 

The time-varying outcome variable $Y_{t+1}$ is generated by a function with the time-varying treatment variable $W_t$,  the conditioning set $\mathbf{Z}_t$ and latent variables $\mathbf{U}_t$:
\begin{equation}
	\begin{aligned}
		Y_{t + 1} = \rho_{{W}}W_t + \rho_{\mathbf{Z}}\mathbf{Z}_{t} + \rho_{\mathbf{U}}\mathbf{U}_{t}, 
	\end{aligned}
\end{equation}where $\rho_{{W}}, \rho_{\mathbf{Z}},\rho_{\mathbf{U}} = 0.5$. 

We generate synthetic time-series datasets, $\mathcal{D}$, with sample sizes of 2k, 4k, 6k, and 8k. To minimise potential bias from the data generation process, we produce 30 datasets for each sample size. Time dependencies are introduced by configuring $p$ to 1 and 3. Both the observed covariates, $\mathbf{X}_t$, and the latent covariates are assigned a dimensionality of 3.

\paragraph{Methods for Comparison}
We evaluate our TDCIV against several state-of-the-art causal effect estimators to highlight its effectiveness. These methods  include:

\begin{enumerate} [leftmargin=0.8cm]
    \item {LinearDML (LDML)}~\cite{chernozhukov2018double}, which addresses reverse causal metric bias using a cross-fitting strategy.
    \item {NonParamDML (NPDML)}~\cite{chernozhukov2018double}, a non-parametric version of Double ML estimators that allows for arbitrary machine learning models in the final stage.
    \item {SparseLinearDML (SLDML)}~\cite{Semenova2023Inference}, an adaptation of LinearDML where the loss function is modified by incorporating $L_1$ regularisation.
    \item {CausalForestDML (CFDML)}~\cite{athey2019generalized}, which employs two random forests to estimate causal effects by predicting two potential outcomes.
    \item {KernelDML (KDML)}~\cite{nie2021quasi}, which integrates dimensionality reduction techniques with kernel methods.
    \item {Meta-learner}~\cite{kunzel2019metalearners}, specifically the X-learner (XLer) and the T-learner (TLer).
\end{enumerate}

\begin{figure*}[t]
	\centering
	\includegraphics[scale=0.425]{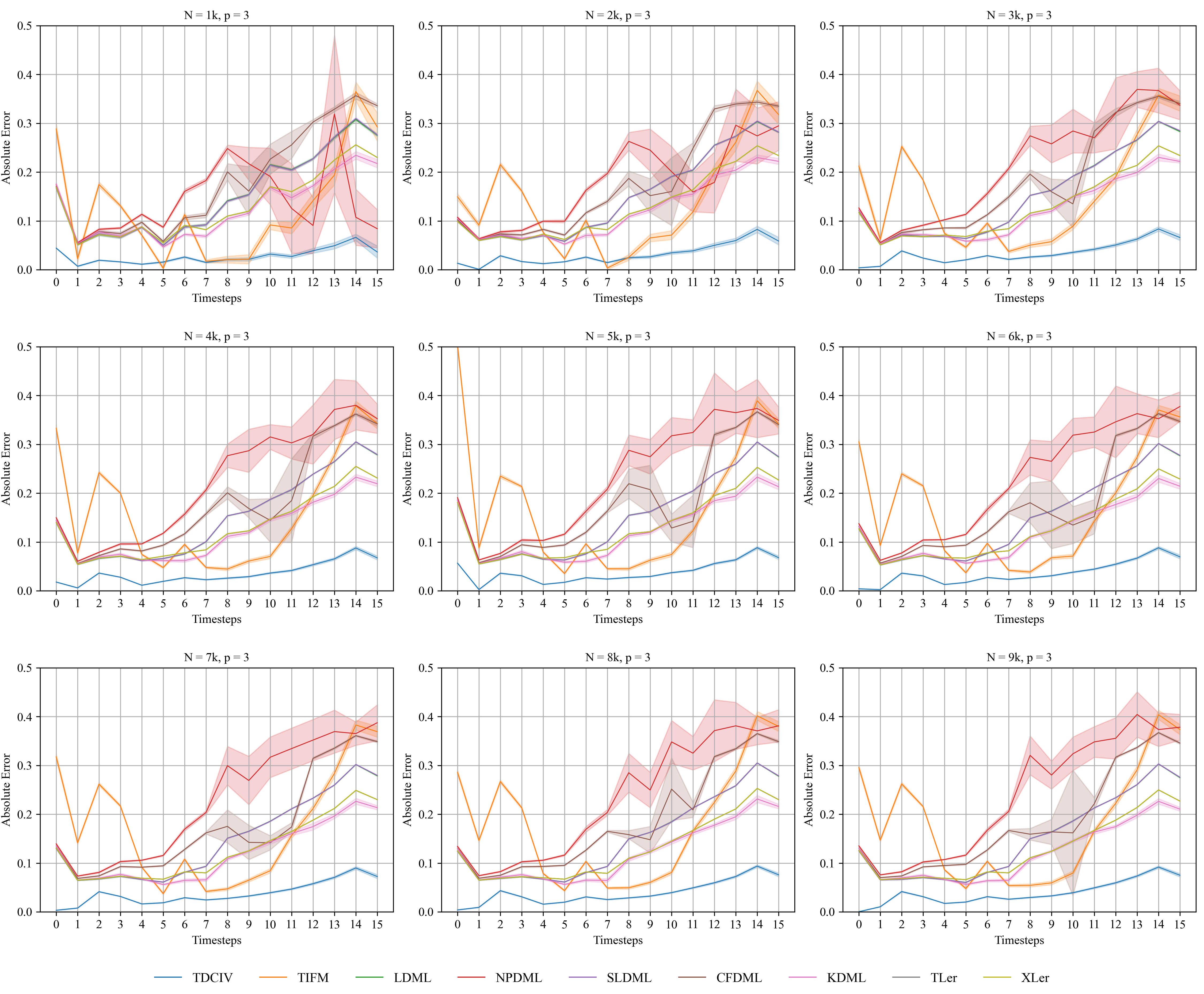}
	\caption{Absolute error results for all methods, presented with their means and standard deviations over 30 synthetic datasets.}
	\label{fig:res_simulate3}
\end{figure*}

It is important to note that methods designed for causal inference in static settings may lead to biased estimates when applied to time-series data, as they fail to account for the temporal dependencies between covariates. In our experiments, we do not compare with IV-based or CIV-based methods developed for static settings, as these approaches assume the IV or CIV is explicitly known and included in the dataset. However, in our problem setting, the IV or CIV is unmeasured and instead proxied by observed variables.

Several methods exist for estimating time-varying causal effects in time-series data, including Standard Marginal Structural Models~\cite{robins2000marginal, hernan2001marginal}, Recurrent Marginal Structural Networks~\cite{lim2018forecasting}, and time-series Deconfounder~\cite{bica2020time}. However, these methods are completely different from the tasks we focus on; they produce predictions of potential outcomes, while our TDCIV method focuses on the estimation of $\hat{ACE}_t$ for each time step. 

There are two works that are similar to our setting. CIV.VAE~\cite{cheng2023causal} learns the representation of CIV and its conditioning set in a static setting. We do not include this method as  TDCIV works with time-series data. TIFM~\cite{cheng2024instrumental} is proposed for handling time-varying latent confounders in time-series data using standard IV. We involve this method as a comparison in our experiment. However, please note that TIFM works with IV rather than CIV, which imposes a more restrictive assumption compared to TDCIV.

\paragraph{Evaluation Criterion}
To assess the performance of TDCIV and the baseline models, we employ the absolute error, $|\hat{ACE_t} - ACE_t|$ as the evaluation metric, where $\hat{ACE_t}$ represents the estimated value, and $ACE_t$ denotes the ground truth.

\paragraph{Implementations}
The compared methods are implemented using the $Python$ package $econml$~\cite{battocchi2019econml}, while our proposed TDCIV method is implemented with TensorFlow~\cite{abadi2016tensorflow}. The parameter configurations for TDCIV are summarised in Table~\ref{tab:setting}.  The source code will be made publicly accessible upon publication of this paper.

\begin{table}[t]
	\centering
	\caption{Parameter settings involved  in our TDCIV method.}
	\label{tab:setting}
	\begin{tabular}{|cc|cc|}
		\toprule
		Parameters & Value & Parameters & Value  \\ \midrule
		Reps & 30 & RNN hidden units & 128  \\
		Epoch & 100 & FC hidden units & 128 \\
		Batch\_Size & 128 & Dropout probability & 0.8  \\ \bottomrule
	\end{tabular}
\end{table}

\begin{table*}[t]
	\centering
	\caption{The performance of the proposed TDCIV and comparison methods is evaluated across varying dimensions of measured covariates. The first column represents the number of measured covariates, while the second column denotes the number of time steps. The best results are highlighted for clarity.}
	\label{tab:setting02}
    \begin{tabular}{ccccccccccc}
    \toprule\midrule
                       &         & TDCIV                  & TIFM          & LDML          & NPDML         & SLDML         & CFDML         & KDML          & TLer      & XLer      \\ \midrule\midrule
    \multirow{4}{*}{3} & t-1  & \textbf{0.032±0.001} & 0.286±0.001 & 0.351±0.002 & 0.423±0.051 & 0.352±0.002 & 0.332±0.072 & 0.339±0.021 & 0.281±0.000 & 0.281±0.000 \\
                       & t-5  & \textbf{0.023±0.002} & 0.091±0.001 & 0.222±0.002 & 0.253±0.005 & 0.223±0.004 & 0.196±0.002 & 0.200±0.007   & 0.179±0.000 & 0.179±0.000 \\
                       & t-10 & \textbf{0.018±0.001} & 0.058±0.001 & 0.190±0.001  & 0.205±0.003 & 0.189±0.000     & 0.161±0.004 & 0.140±0.006  & 0.130±0.000  & 0.130±0.000  \\
                       & t-15 & \textbf{0.017±0.001} & 0.070±0.001  & 0.195±0.001 & 0.219±0.003 & 0.195±0.001 & 0.150±0.002  & 0.141±0.006 & 0.139±0.000 & 0.139±0.000 \\ \midrule
    \multirow{4}{*}{6} & t-1  & \textbf{0.037±0.001} & 0.321±0.001 & 0.431±0.002 & 0.441±0.045 & 0.431±0.002 & 0.329±0.064 & 0.338±0.025 & 0.283±0.000 & 0.283±0.000 \\
                       & t-5  & \textbf{0.024±0.004} & 0.132±0.001 & 0.307±0.017 & 0.297±0.015 & 0.300±0.005   & 0.254±0.015 & 0.267±0.026 & 0.204±0.000 & 0.204±0.000 \\
                       & t-10 & \textbf{0.015±0.001} & 0.061±0.001 & 0.204±0.001 & 0.226±0.003 & 0.203±0.001 & 0.177±0.006 & 0.155±0.006 & 0.140±0.000  & 0.140±0.000  \\
                       & t-15 & \textbf{0.016±0.001} & 0.068±0.001 & 0.208±0.001 & 0.218±0.004 & 0.209±0.003 & 0.177±0.004 & 0.170±0.007  & 0.149±0.000 & 0.149±0.000 \\ \midrule
    \multirow{4}{*}{9} & t-1  & \textbf{0.039±0.001} & 0.324±0.001 & 0.433±0.008 & 0.453±0.032 & 0.434±0.008 & 0.355±0.088 & 0.362±0.015 & 0.297±0.000 & 0.297±0.000 \\
                       & t-5  & \textbf{0.022±0.002} & 0.195±0.001 & 0.335±0.022 & 0.322±0.019 & 0.334±0.023 & 0.276±0.029 & 0.295±0.025 & 0.246±0.000 & 0.246±0.000 \\
                       & t-10 & \textbf{0.016±0.001} & 0.078±0.001 & 0.215±0.002 & 0.225±0.005 & 0.213±0.001 & 0.188±0.004 & 0.179±0.006 & 0.161±0.000 & 0.161±0.000 \\
                       & t-15 & \textbf{0.017±0.003} & 0.089±0.001 & 0.308±0.003 & 0.398±0.063 & 0.307±0.002 & 0.225±0.005 & 0.314±0.015 & 0.289±0.000 & 0.289±0.000 \\ \midrule\bottomrule
    \end{tabular}
\end{table*}

\begin{table*}[t]
	\centering
	\caption{The performance of the proposed TDCIV and comparison methods is evaluated across varying dimensions of unmeasured covariates. The first column represents the number of unmeasured covariates, while the second column denotes the number of time steps. The best results are highlighted for ease of interpretation.}
	\label{tab:setting03}
    \begin{tabular}{ccccccccccc}
    \toprule\midrule
                       &         & TDCIV                    & TIFM          & LDML          & NPDML         & SLDML         & CFDML         & KDML          & TLer          & XLer          \\ \midrule\midrule
    \multirow{4}{*}{3} & t-1  & \textbf{0.038±0.001} & 0.322±0.001 & 0.351±0.002 & 0.423±0.051 & 0.352±0.002 & 0.355±0.088 & 0.338±0.025 & 0.281±0.000 & 0.281±0.000 \\
                       & t-5  & \textbf{0.039±0.002}   & 0.230±0.001 & 0.381±0.002 & 0.412±0.036 & 0.383±0.008 & 0.341±0.015 & 0.298±0.010 & 0.240±0.000 & 0.240±0.000 \\
                       & t-10 & \textbf{0.028±0.002}   & 0.217±0.001 & 0.402±0.026 & 0.409±0.037 & 0.400±0.024 & 0.364±0.029 & 0.351±0.040 & 0.258±0.000 & 0.258±0.000 \\
                       & t-15 & \textbf{0.024±0.003}   & 0.132±0.001 & 0.307±0.017 & 0.297±0.015 & 0.300±0.005 & 0.254±0.015 & 0.267±0.026 & 0.204±0.000 & 0.204±0.000 \\ \midrule
    \multirow{4}{*}{6} & t-1  & \textbf{0.039±0.001} & 0.361±0.001 & 2.852±0.007 & 2.57±0.051  & 1.85±0.007  & 4.301±0.071 & 1.886±0.011 & 1.933±0.000 & 1.933±0.000 \\
                       & t-5  & \textbf{0.028±0.002}   & 0.249±0.001 & 1.939±0.006 & 1.887±0.020 & 1.338±0.006 & 3.216±0.014 & 1.698±0.009 & 1.708±0.000 & 1.708±0.000 \\
                       & t-10 & \textbf{0.021±0.002}   & 0.226±0.001 & 1.602±0.005 & 1.609±0.055 & 0.893±0.006 & 3.261±0.005 & 1.638±0.003 & 1.685±0.000 & 1.685±0.000 \\
                       & t-15 & \textbf{0.019±0.002}   & 0.197±0.001 & 0.909±0.001 & 1.383±0.025 & 0.712±0.001 & 2.417±0.097 & 1.438±0.006 & 1.402±0.000 & 1.402±0.000 \\ \midrule
    \multirow{4}{*}{9} & t-1  & \textbf{0.040±0.001} & 0.376±0.001 & 4.949±0.003 & 5.008±0.037 & 4.951±0.004 & 4.512±0.082 & 4.928±0.019 & 4.947±0.000 & 4.947±0.000 \\
                       & t-5  & \textbf{0.032±0.001}   & 0.289±0.001 & 3.319±0.002 & 3.284±0.021 & 3.319±0.003 & 3.015±0.077 & 3.334±0.022 & 3.273±0.000 & 3.273±0.000 \\
                       & t-10 & \textbf{0.025±0.002}   & 0.177±0.001 & 3.160±0.009 & 2.706±0.091 & 3.162±0.011 & 5.024±0.010 & 3.164±0.021 & 3.161±0.000 & 3.161±0.000 \\
                       & t-15 & \textbf{0.021±0.001}   & 0.162±0.001 & 2.553±0.027 & 2.280±0.086 & 2.579±0.031 & 3.477±0.018 & 2.471±0.020  & 2.468±0.000 & 2.468±0.000 \\ \midrule\bottomrule
    \end{tabular}
\end{table*}

\subsection{Evaluation on Synthetic Datasets}
The absolute errors of all methods across 30 synthetic datasets are presented in Figure~\ref{fig:res_simulate} and Figure~\ref{fig:res_simulate3}. These results demonstrate that the TDCIV method consistently achieves the lowest absolute error compared to other approaches. While the baselines account for covariates to address bias in static settings, they struggle to handle latent and time-dependent confounders, leading to higher estimation errors. Although TIFM outperforms other comparison methods in static scenarios, it still falls short of TDCIV. This result aligns with expectations, as no valid standard IV exists in this setup. Consequently, TIFM may inadvertently introduce bias by learning an inaccurate substitute for the standard IV, affecting causal effect estimation.

Our TDCIV method addresses these challenges by learning the time-varying causal influence variable (CIV) $\bar{S}_t$ from proxy variables and generating the conditioning set $\bar{\mathbf{Z}}_t$ based on the observed covariates $\bar{\mathbf{X}}_t$. The CIV-based estimator is then used to estimate $ACE(w_t, Y_{t+1})$. This approach enables TDCIV to effectively handle bias arising from time dependencies, while the CIV estimator mitigates bias caused by time-varying  latent confounders. As a result, TDCIV consistently outperforms other methods. However, we note that there is a high error value at the first time step in some sample sizes, as the first time step is initialised with random values, which limits our ability to guarantee performance at that point.

\subsection{Sensitivity Analysis}
We conduct a sensitivity analysis to examine the impact of the number of measured covariates, $\mathbf{X}_{t}$, on the performance of  our TDCIV  and other compared models. The experimental setting is fixed with a sample size of 5k, while varying the number of measured covariates, $\mathbf{X}_{t}$. Table~\ref{tab:setting02} presents the comparative results under different numbers of measured covariates, and we observe that the proposed TDCIV consistently performs the best compared to all other models. Specifically, there are no significant changes in performance as the number of measured covariates increases. This trend aligns with our expectations, as a higher number of measured covariates implies increased complexity of observed confounding bias. However, all models are able to handle this bias effectively.

Furthermore, we conduct a sensitivity analysis on the number of unmeasured covariates, $\mathbf{U}_{t}$, to explore its impact on the performance of our proposed TDCIV and other comparison models. Table~\ref{tab:setting03} shows the results under different numbers of unmeasured covariates. We observe that the proposed TDCIV consistently outperforms all other comparison models. However, there is a general decrease in performance among the comparison models, except for TIFM. Notably, only IV-based methods (i.e., the proposed TDCIV and TIFM) maintain stable performance regardless of the increasing number of unmeasured covariates, as IV-based methods can handle the presence of unmeasured covariates. Specifically, TDCIV outperforms TIFM, demonstrating that the CIV is easier to satisfy in complex scenarios.

\subsection{Case Study on Climate Change Data}
The dataset used in our work is sourced from the National Centers for Environmental Prediction (NCEP) and the National Center for Atmospheric Research (NCAR). This dataset, a cornerstone in atmospheric research~\cite{kalnay1996ncep}, includes a wide array of variables such as precipitation rate (prate), pressure level (pres), air temperature (air), skin temperature (skt), downward shortwave radiation flux (dswrf), clear-sky upward solar flux (csusf), clear-sky downward longwave flux (csdlf), cloud forcing net longwave flux (cfnlf), wind speed (wspd), minimum temperature (tmin), and seasonal categories (season).

For this analysis, we concentrate on two regions: Europe (418 data points) and the United States (403 data points). The treatments of interest, cfnlf, wspd, and skt, are analysed to determine their causal effects on prate at each time step. The study covers 120 monthly intervals from 2013 to 2022. 

\begin{figure}[t]
	\centering
	\includegraphics[scale=0.345]{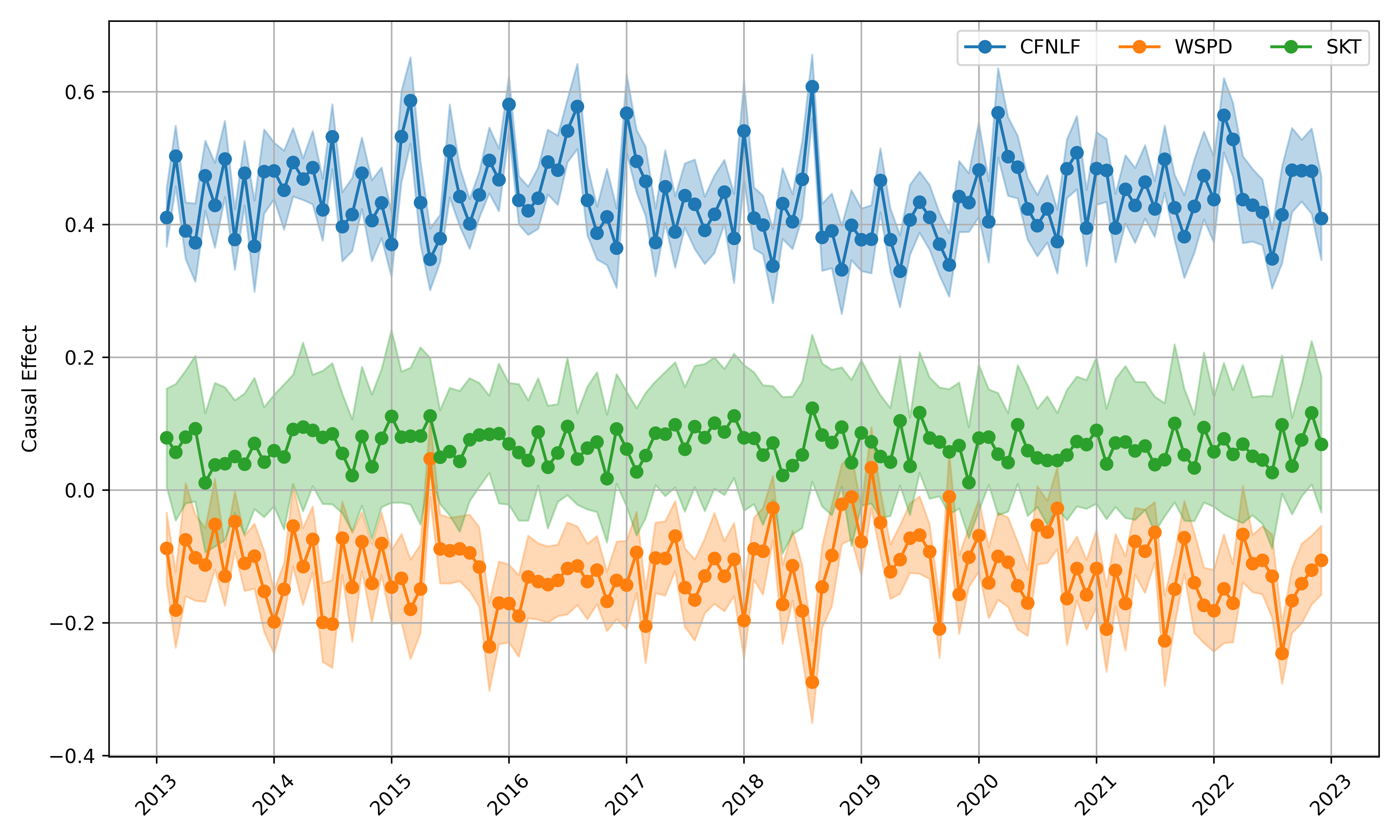}
	\caption{The time-varying causal effects estimated by our proposed TDCIV method on the Europe dataset.}
	\label{pic:europe}
\end{figure}

\begin{figure}[t]
	\centering
	\includegraphics[scale=0.345]{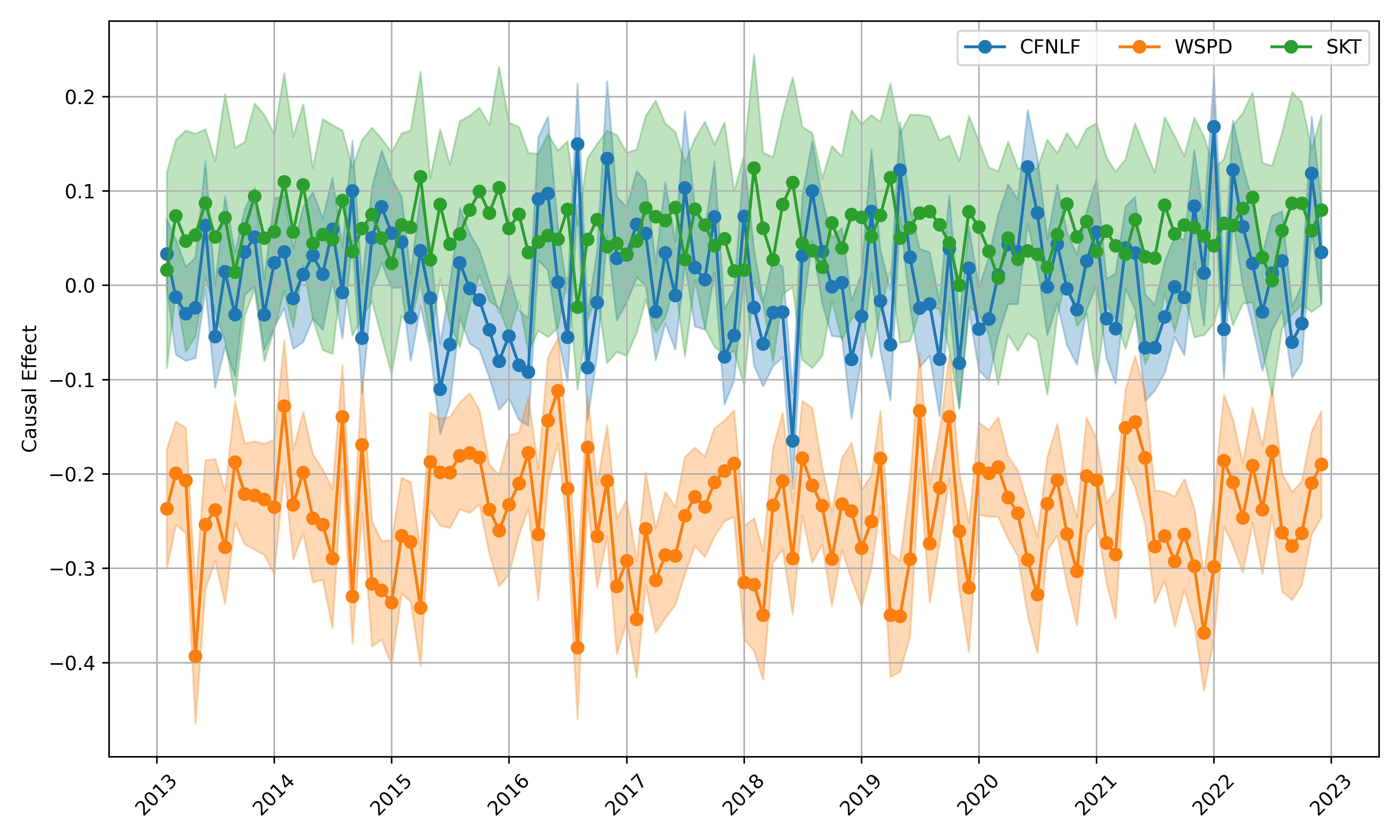}
	\caption{The time-varying causal effects estimated by our proposed TDCIV method on the United States dataset.}
	\label{pic:us}
\end{figure}

Figure~\ref{pic:europe} illustrates the results for Europe, where wind speed indirectly influences rainfall by affecting cloud formation and distribution through water vapor transport. Rainfall occurs when clouds accumulate sufficiently large water droplets~\cite{haylock2008european, uppala2005era, zolina2013changes}. The analysis confirms that cfnlf has the most substantial causal effect on prate in Europe, consistent with domain knowledge.
In contrast, Figure~\ref{pic:us} shows that skin temperature (skt) is the primary driver of rainfall in the United States. This result aligns with expert knowledge, as the country’s varied climatic conditions make surface temperature a key determinant of rainfall, particularly in regions where high temperatures trigger strong convective activity~\cite{seeleyromps2020}.

The proposed TDCIV method effectively captures the interactions between latent factors and historical records. By leveraging historical records as proxies for CIVs, the experimental findings confirm TDCIV’s effectiveness in tackling real-world causal inference challenges.

\section{Related work}
\label{sec:relatedwork}
In this section, we reviewed works  related to our TDCIV, including methods using static IVs, methods for causal effect estimation in time-series data, and methods leveraging time-varying IVs. 

\paragraph{Methods using static IVs}
Most existing methods for causal effect estimation from observational data assume that all confounders, i.e., variables that affect both the treatment variable and the outcome variable, are static~\cite{imbens2015causal}, meaning they do not vary over time. Numerous approaches have been developed to estimate causal effects in static environments, where all variables are assumed to remain constant~\cite{shalit2017estimating, louizos2017causal, zhang2021treatment, cheng2023causal, xu2023disentangled, XuCLL0Y24, ChengXL0LL24}. However, this assumption is \textit{unrealistic} in many real-world scenarios and limits their ability to discover and utilise causal relationships in temporal settings.

The IV approach is a valuable tool for addressing confounding bias caused by latent confounders in causal inference using observational data ~\cite{hernan2006instruments,imbens2014instrumental}. Most existing IV-based methods operate in static data, where the treatment and outcome are assumed to keep constant over time~\cite{hernan2006instruments}. One of the most commonly used IV methods is the two-stage least squares (2SLS)~\cite{angrist1995two}. Recently, Hartford et al.~\cite{hartford2017deep} developed a deep learning-based IV method, DeepIV, which utilises deep representation learning to estimate causal effects in nonlinear settings. Athey et al.~\cite{athey2019generalized} extended the causal random forest to develop an IV method, i.e., the generalised random forest method. Unlike these methods that rely on static IVs, our work aims to develop a novel time-varying IV approach for causal effect estimation in time-series data with latent confounders.
 
\paragraph{Methods for causal effect estimation without IV-Based Approaches}
Traditional methods for causal effect estimation in time-series data were primarily developed in epidemiology to address time-varying outcomes. These include approaches such as $g$-computation, structural nested mean models (SNMMs), and marginal structural models (MSMs)~\cite{robins1986new,robins1997causal,robins2000marginal}. However, these methods often rely on logistic or linear regression models, limiting their ability to capture complex, time-varying relationships.

To overcome these limitations related to estimating time-varying causal effects in time-series data,  Lim et al.~\cite{lim2018forecasting} developed recurrent marginal structural networks (RMSNs), which predict these effects by incorporating temporal dependencies.  Bica et al.~\cite{bica2019estimating} introduced the counterfactual recurrent network (CRN), designed to construct treatment-invariant representations that mitigate confounding bias arising from time-varying confounders. Building on these approaches, Bica et al. further proposed the time-series deconfounder (TSD)~\cite{bica2020time}, leveraging recurrent neural networks (RNNs) to estimate causal effects across multiple treatments over time. 

Additionally, Melnychuk et al.~\cite{melnychuk2022causal} introduced the Causal Transformer (CT) model, designed to capture complex, long-range dependencies among time-varying confounders in observational data. Sun et al.~\cite{sun2023CPT} developed the Causal Trajectory Prediction (CTP) model, which combines trajectory prediction with causal discovery to forecast the progression of non-communicable diseases, enhancing interpretability in clinical decision-making. Similarly, Cao et al.~\cite{cao2023estimating} presented LipCDE, which uses Lipschitz regularisation and neural controlled differential equations to reduce bias caused by time-varying latent confounders. Frauen et al.~\cite{frauen2023estimating} developed DeepACE, using the iterative $g$-computation formula to estimate time-varying causal effects while effectively addressing the impact of latent confounders.

Despite these advances, many existing methods encounter challenges when addressing latent time-dependent confounders in single treatment-outcome caes. For instance, models such as TSD and LipCDE are developed for multiple time-varying treatments, whereas methods like DeepACE and CT are not equipped to handle latent confounders. 

In contrast, our TDCIV  specifically addresses the challenge of latent time-dependent confounders in single treatment-outcome relationships by developing a novel CIV-based approach for time-series data.

\paragraph{Methods leveraging time-varying IVs}
Recently, several time-varying IV methods have been developed for estimating time-varying causal effects, particularly in the presence of time-to-event outcomes with latent time-varying confounders. For instance, Martinussen et al.~\cite{martinussen2017instrumental} proposed a time-varying IV estimator within a semiparametric structural cumulative model, utilising a structural accelerated failure model. Furthermore, Michael et al.~\cite{michael2023instrumental} explored the identification and estimation of marginal structural mean models (MSMMs) for time-varying treatments, incorporating the use of time-varying IVs. Cui et al.~\cite{cui2023instrumental} extended the work of Michael et al. by providing sufficient conditions for parameter identification in MSMs using temporal data, leveraging time-varying IVs derived from domain knowledge. However, these time-varying IV-based methods rely on external domain expertise to define a valid time-varying IV. 

In contrast, our TDCIV aims to disentangle and learn the representations of time-varying CIV and its conditioning set  from time-series data, thereby minimising reliance on pre-existing domain knowledge.
 
\section{Conclusion}
\label{sec:con}
In our paper, we develop a novel  debiased method, TDCIV, for learning representations of a time-varying CIV and its conditioning set to estimate causal effect in time-series data. TDCIV aims to address confounding bias caused by time-varying latent confounders. To achieve this, we generalise the concept of CIV in causal DAG from static to time-varying settings. Our TDCIV integrates VAEs with LSTM models to capture the complex temporal relationships between latent factors and measured variables in time-series data. By disentangling and reconstructing the latent representations ${S}_t$ and ${\mathbf{Z}}_t$, and integrating them with historical data, TDCIV effectively mitigates the confounding bias introduced by time-varying latent confounders.  Experimental evaluations on both synthetic and real-world datasets validate the effectiveness of TDCIV in estimating causal effects in time-series data influenced by latent confounders. In future work, we aim to explore broader applications of the time-varying CIV in more complex scenarios and extend its utility to address additional challenges in time-series causal inference.

\bibliographystyle{IEEEtran}
\bibliography{TDCIV.bib}

\end{document}